\documentclass[letterpaper]{article} 
\usepackage{aaai2026}  
\usepackage{times}  
\usepackage{helvet}  
\usepackage{courier}  
\usepackage[hyphens]{url}  
\usepackage{graphicx} 
\usepackage{algorithm}
\usepackage{algorithmic}
\usepackage[table]{xcolor} 
\usepackage{multirow}
\usepackage{amsfonts} 
\usepackage{natbib}  
\usepackage{caption} 
\usepackage{newfloat}
\usepackage{listings}
\DeclareCaptionStyle{ruled}{labelfont=normalfont,labelsep=colon,strut=off} 
\floatstyle{ruled}
\newfloat{listing}{tb}{lst}{}
\floatname{listing}{Listing}
\nocopyright

\title{Adversarial Attacks on Robot Localization Systems via Deep Feature Perturbation}
\author{
    Zhenyu Li$^{1, *}$, Tianyi Shang$^2$
}
\affiliations{
    \textsuperscript{\rm 1}Shandong Academy of Sciences\\
    \textsuperscript{\rm 2}Fuzhou University\\

}

\usepackage{bibentry}

\begin{document}

\maketitle

\begin{abstract}
Robot localization systems are critical for autonomous navigation and safety. Adversarial perturbations can mislead these systems, resulting in mislocalization, navigation errors, or unsafe interactions, especially in mission-critical scenarios. This paper investigates the vulnerability of deep learning–based localization pipelines to adversarial attacks. We propose a novel framework for generating adversarial queries that specifically target Product Quantization (PQ) in visual localization systems. Our method employs a Lightweight Product Quantization Network (LPQN) to perturb query feature encodings, misleading the retrieval process by returning semantically irrelevant database entries. Adversarial queries are generated via a two-phase procedure: a forward pass that perturbs feature distributions and a backward pass that refines the perturbation through optimization. The lightweight design of LPQN allows creation of subtle yet highly effective perturbations with minimal computational overhead. Extensive experiments in both controlled and real-world robotic environments demonstrate that our approach substantially degrades PQN performance, exposing critical vulnerabilities in practical applications.
\end{abstract}


\section{Introduction}
\begin{figure}
    \centering
    \includegraphics[width=0.95\linewidth]{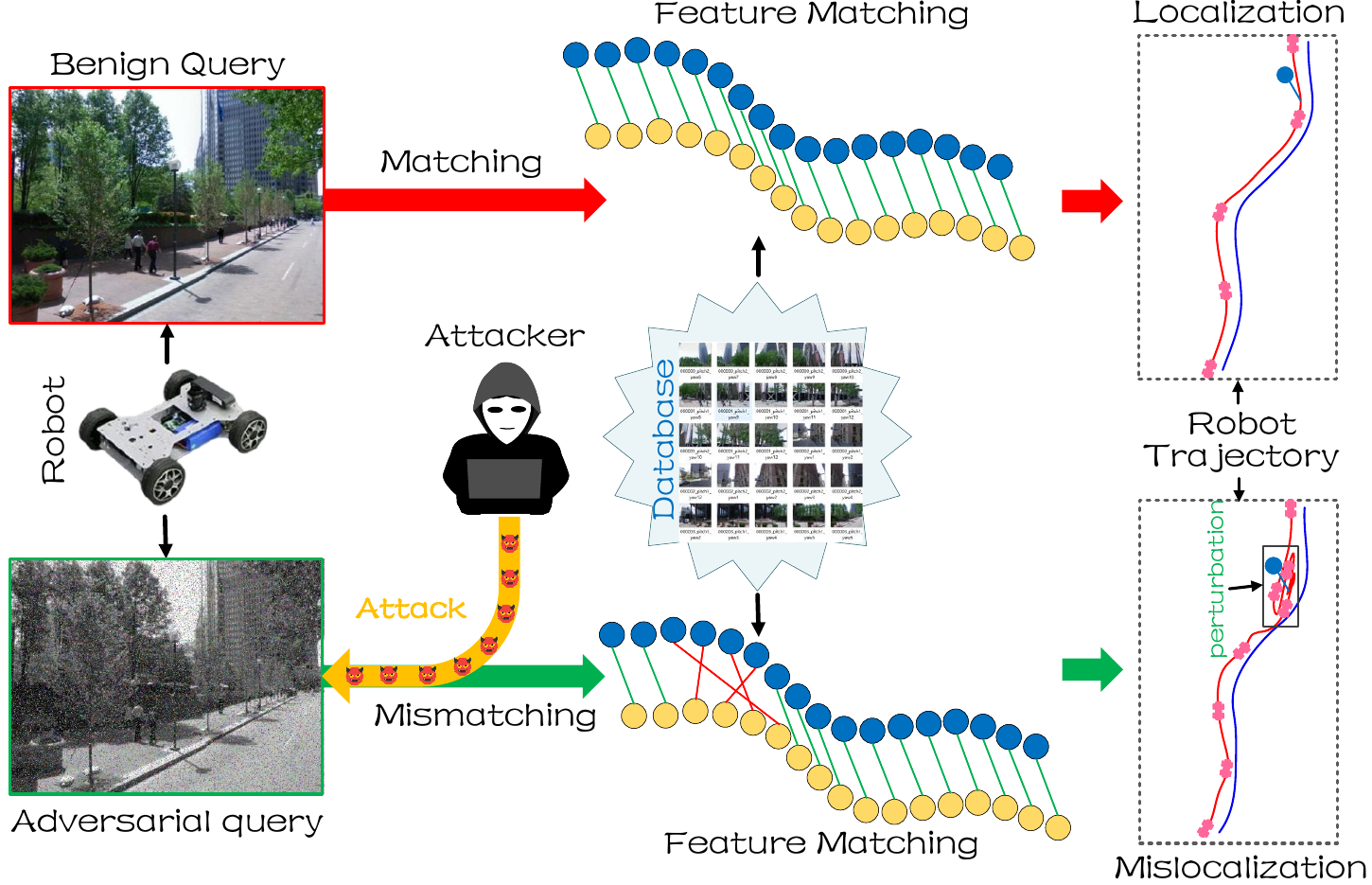}
    \caption{An example of an adversarial attack against a robot localization system. By adding perturbations to the feature space, a mismatch was successfully induced between the query and the reference database, ultimately resulting in the failure of localization.}
    \label{fig1}
\end{figure}

Robot localization is a fundamental aspect of autonomous systems, enabling robots to navigate and interact accurately with their environments \cite{zhang2022variational}, \cite{lee2022improved}. Localization systems often rely on place recognition methods \cite{li2025place}, using product quantization networks (PQNs) \cite{rokh2023comprehensive}, which enable robots to estimate their place by matching visual inputs to a predefined set of environmental features. These systems are essential for a wide range of applications, including autonomous vehicles, industrial robots, healthcare, and service robots. An example of an adversarial attack against a robot localization system is shown in Figure \ref{fig1}.
Recent research has demonstrated that adversarial attacks on deep learning models, particularly in visual perception tasks, can degrade system performance by subtly altering the model's behavior with imperceptible perturbations. Despite the critical role of localization in autonomous robotics, the impact of adversarial attacks on localization systems that utilize PQNs remains underexplored. Existing adversarial machine learning methods primarily focus on defending against attacks in classification or detection tasks, leaving significant gaps in addressing the specific vulnerabilities of localization pipelines. For instance, traditional defenses such as adversarial training and robust optimization techniques have been shown to improve robustness in image classification tasks but often fail to generalize to retrieval-based localization systems, where the problem involves continuous vector quantization rather than discrete classification labels.

Compared to previous work on adversarial attacks targeting retrieval-based systems, such as those focused on image hashing or other quantization techniques \cite{chen2022adversarial}, \cite{tang2024once}, our proposed method introduces a Lightweight Product Quality Network (LPQN) specifically designed for robot localization tasks. Our approach leverages a novel adversarial query generation framework that directly perturbs the centroid assignment probability distributions within the quantization space. This method differs from traditional adversarial techniques, which typically rely on directly modifying image features or using gradient-based methods for classification models. By exploiting vulnerabilities in the soft product quantization layer, our LPQN ensures that perturbations are small yet impactful, causing significant mislocalization without incurring high computational costs.

While prior works such as NetVLAD and CosPlace focus on feature aggregation and robustness to viewpoint changes, none explicitly consider vulnerabilities introduced by product quantization. Our approach fills this gap by directly targeting PQ centroid assignments, exposing a previously unexplored attack surface. The key advantages of our method lie in its efficiency and scalability. Unlike existing approaches, which often entail high computational costs or require retraining models, our LPQN framework is lightweight and can be applied to pre-trained PQN systems without substantial overhead. Furthermore, our approach addresses specific weaknesses in product quantization networks, particularly the non-differentiable nature of the quantization operation, by providing a differentiable approximation that enables effective adversarial generation. By leveraging this approximation, we can craft adversarial queries that significantly disrupt localization accuracy, even in real-world robotic environments, without necessitating complex modifications to the underlying model architecture. The main contributions of this paper are as follows: 1) We introduce a novel adversarial attack framework for robot localization systems based on PQNs. Our method targets centroid assignment distributions to generate effective perturbations, overcoming the indifferentiability of product quantization layers. 2) We propose a lightweight LPQN with residual connections that enables the generation of adversarial queries with reduced computational overhead. The network can output descriptors of arbitrary length, enhancing flexibility while maintaining attack effectiveness. 3) The proposed method is validated on real-world robotic localization systems, demonstrating significant performance degradation in PQN-based systems, highlighting the importance of robust security in robotic navigation.
\section{Related Work}
\subsection{Visual Localization for Robots}
Visual localization is a fundamental component of robot loop closure detection and long-term autonomous navigation \cite{li2024feature}, \cite{li2025multi}. Its goal is to determine whether a robot has previously visited a location using visual observations \cite{li2024toward}. Recent surveys indicate that visual localization methods have evolved from handcrafted pipelines to deep learning–based global retrieval frameworks, reflecting the growing demand for both robustness and scalability in real-world robotic deployments. 

A major milestone in visual localization is NetVLAD \cite{rokh2023comprehensive}, which introduced a differentiable VLAD aggregation layer and established the standard paradigm of learning global place descriptors in an end-to-end manner. Building on this foundation, subsequent methods have focused on improving descriptor robustness and retrieval accuracy under severe appearance and viewpoint changes. For example, PatchNetVLAD \cite{hausler2021patch} enhanced global retrieval by incorporating patch-level matching, enabling finer-grained place correspondence and improving robustness in challenging scenarios. TransVPR \cite{wang2022transvpr} further integrated Transformer-based modeling to better capture global contextual dependencies, reflecting the trend toward stronger long-range feature interactions in visual localization.

Another important research direction is discriminative representation learning for large-scale place retrieval. CosPlace \cite{berton2022rethinking} showed that explicitly learning cosine-based discriminative embeddings can significantly improve localization performance while maintaining training efficiency. EigenPlaces \cite{berton2023eigenplaces} further advanced this direction by learning more generalizable place representations with strong discrimination ability across environments. More recently, MixVPR \cite{ali2023mixvpr} proposed a simple yet effective feature mixer for visual localization, demonstrating that lightweight architectural design can also yield competitive retrieval performance. Similarly, SelaVPR \cite{lu2024towards} and CricaVPR \cite{lu2024cricavpr} explored improved feature aggregation and contextual reasoning strategies to enhance descriptor quality, particularly in complex urban or large-scale scenes.

In addition to improving localization accuracy, efficiency, and compactness have become increasingly important in robotic deployment. Since onboard localization systems are often constrained by storage, computation, and latency, recent methods have paid greater attention to compact yet expressive descriptors. SALAD \cite{izquierdo2024optimal} and $A^2$GC \cite{li20252} represent this trend by designing more efficient global aggregation strategies with strong retrieval performance. Very recent works such as FoL \cite{wang2025focus} and Pair-VPR \cite{hausler2025pair} further highlight the ongoing effort to improve localization accuracy while preserving deployment efficiency. \textit{Despite these advances, current visual localization methods still exhibit significant limitations for robotic localization. Most approaches primarily focus on optimizing retrieval accuracy, while giving limited consideration to adversarial robustness and security during robot deployment.}
\subsection{Adversarial Attack for Visual Tasks}
Adversarial attacks have revealed a fundamental vulnerability in deep neural networks (DNNs) across a wide range of visual tasks \cite{zhou2022adversarial}. By introducing carefully crafted and often imperceptible perturbations to the input, attackers can cause deep models to generate highly unreliable predictions. Since the seminal studies on adversarial examples in image classification, adversarial machine learning has rapidly expanded to encompass broader computer vision applications.

Adversarial attacks on image retrieval differ fundamentally from classification attacks: instead of changing a discrete label, they aim to corrupt the ranking structure of the feature space so that relevant images are pushed away, and irrelevant or attacker-chosen images are pulled closer \cite{liu2025robust}. Early research, such as \cite{li2019universal}, established this direction by demonstrating that even a single universal perturbation can systematically disrupt neighborhood relationships and significantly degrade retrieval performance across numerous queries. Subsequent research, such as \cite{tolias2019targeted}, expanded the threat model from untargeted degradation to targeted mismatch attacks. In these attacks, an adversarial query is optimized to appear visually unrelated to the original content while still retrieving nearly identical results, exposing significant privacy and manipulation risks for content-based search engines. 

Universal adversarial perturbations for semantic image segmentation extend the concept of image-agnostic attacks, originally proposed for classification, to structured prediction models that output dense, pixel-level label maps. \cite{croce2023robust} introduces a novel approach to adversarial attacks on semantic segmentation models, highlighting their vulnerability to stronger, pixel-level adversarial perturbations. \cite{rony2023proximal} presents a proximal splitting optimization method for generating effective adversarial attacks on semantic segmentation models, improving attack success and robustness. \cite{rossolini2023real} examines the adversarial robustness of real-time semantic segmentation models for autonomous driving, showing that even robust models are vulnerable to physical and digital adversarial attacks in real-world scenarios.

Visual localization systems offer virtually no protection against carefully crafted attacks. When applied to robot navigation, this vulnerability could lead to severe consequences. \cite{malone2025adversarial} explores adversarial attacks on visual place recognition (VPR) systems for autonomous navigation, proposing detection mechanisms to identify manipulated queries, thereby improving robot safety and ensuring reliable navigation. \cite{claxton2024improving} enhances robot navigation by validating visual localization estimates using consistency verification, improving robustness against false matches, and reducing localization errors to support safer, more reliable autonomous navigation. \cite{ikram2022perceptual} introduces a comprehensive adversarial attack against both front‑end feature matching and back‑end optimization in visual SLAM, exploiting perceptual aliasing to cause severe tracking and mapping failures in autonomous systems. However, compared to our approach, there is a main difference: \textit{our work is the first attempt to study the adversarial generation mechanism in visual localization based on product quantification, while previous research has only focused on adversarial perturbation generation based on hash and attacks on samples.}
\section{Methodology}
\begin{figure*}
    \centering
    \includegraphics[width=0.90\linewidth]{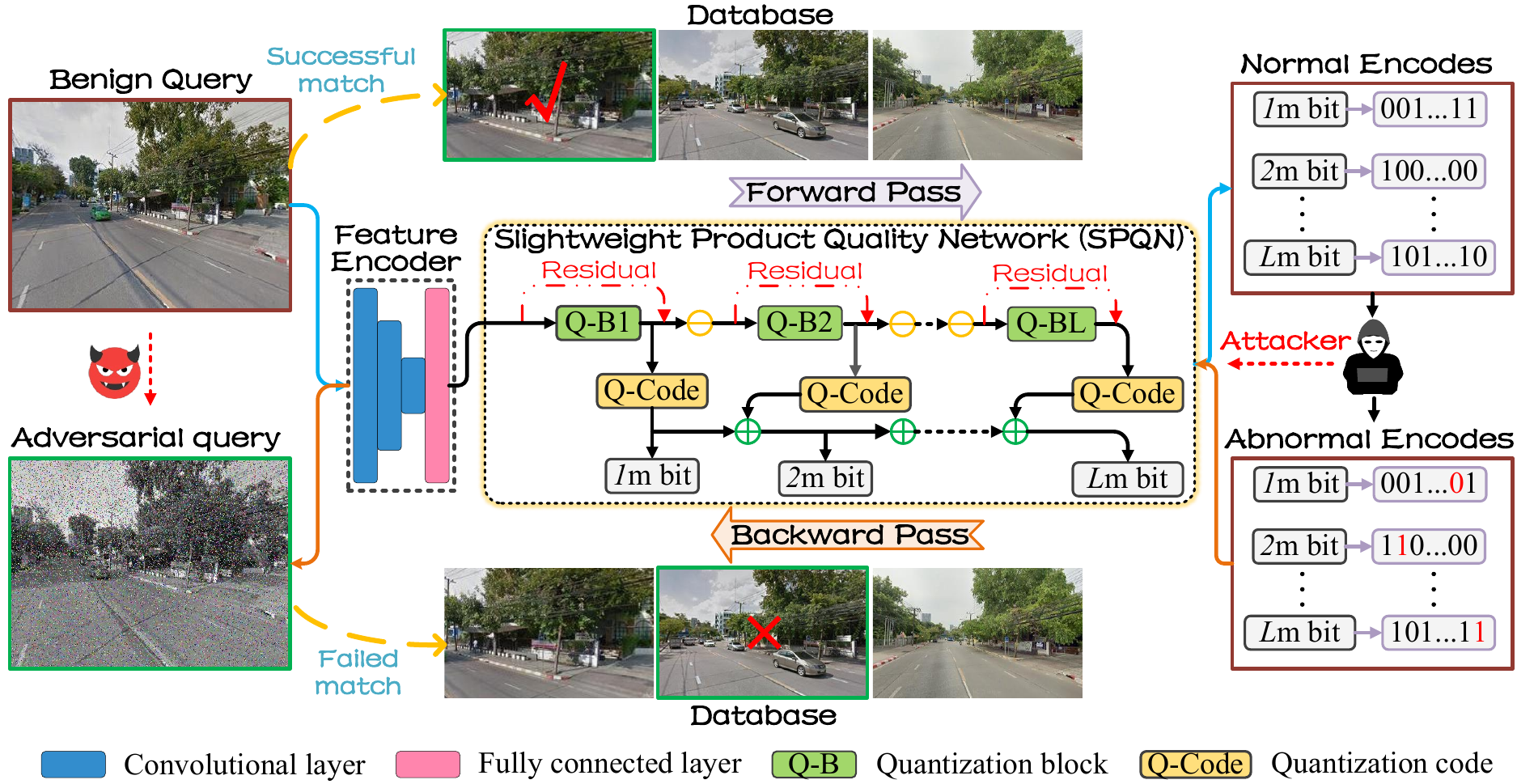}
    \caption{Overview of the Proposed Framework. The adversarial query generation proceeds in two phases: For the forward pass, the feature representation of the benign query is propagated through the LPQN, which injects residual perturbations at each quantization stage to corrupt the intermediate code assignments. For the backward pass, the perturbed encodings are backpropagated through the optimization objective, thereby refining $\delta$ to maximize retrieval failure while satisfying the imperceptibility constraint.}
    \label{fig2}
\end{figure*}

As illustrated in Figure \ref{fig2}, the proposed framework consists of two principal components: 1) a Feature Encoder that extracts high-dimensional visual representations from the query image, and 2) a Lightweight Product Quantization Network (LPQN) that progressively perturbs the quantization codes to mislead the downstream retrieval process. From a theoretical perspective, the vulnerability of PQ-based visual localization originates from the discrete nature of centroid partitioning. The quantization process inherently creates a limited codebook, which can be exploited to construct adversarial perturbations that shift feature embeddings across centroid boundaries, leading to maximum retrieval error.
\subsection{Problem Formulation}
Let $\mathcal{D}=\{d_i\}^{N}_{i=1}$ denote a database of $N$ reference images used for visual localization. Given a query image $q$, captured by the onboard camera, a standard visual localization pipeline first extracts a high-dimensional feature descriptor via a deep convolutional neural network $\phi$: $\chi \rightarrow \mathbb{R}^d$, yielding $f =\phi(q) \in \mathbb{R}^d$. The feature vector is subsequently encoded via PQ, which decomposes the $d$-dimensional space into $L$ disjoint subspaces of equal dimensionality $d/L$. For each subspace $l \in \{1, 2, ...,L\}$, an independent codebook $C^{(l)}=\{c^{(l)}_k\}^{K}_{k=1}$ containing $K$ learned centroids is maintained. The hard quantization operation assigns each sub-vector $f^{(l)} \in \mathbb{R}^{d/L}$ to its nearest centroid:
\begin{equation}
    b^{(l)}=\arg\min_{k\in[K]}\left\|\mathbf{f}^{(l)}-c_k^{(l)}\right\|_2^2,\quad l=1,\ldots,L
\end{equation}
where $[K] = \{1,2,\cdots,K\}$. The resulting $m$-bit code for each subspace is concatenated to form the final compact binary descriptor of length $Lm$ bits:
\begin{equation}
    \mathrm{code}(\mathbf{f})=\bigoplus_{l=1}^{L}\mathrm{onehot}\!\left(b^{(l)}\right)
\end{equation}
where $\oplus$ denotes bit-level concatenation. Database retrieval is performed by computing either Symmetric Distance Computation (SDC) or Asymmetric Distance Computation (ADC) between the query and all database entries in the quantized feature space. The localization result is returned as:
\begin{equation}
    d^{*}=\arg\min_{d_t\subset \mathcal{D}} \mathrm{D_{PQ}}\!\left(\mathbf{f}_q, \mathbf{f}_{d_t}\right)
\end{equation}
where $\mathcal{D} (\cdot,\cdot)$ represents the PQ-based approximate distance. The robot then uses the geographic coordinates of $d*$ as its estimated pose for downstream navigation.

We consider a white-box threat model in which the attacker possesses full knowledge of the feature encoder $\phi$ and the PQ codebooks $C^{(l)}$. Given a benign query image $q$, the attacker seeks to construct an adversarial counterpart $q'=q+\delta$ such that:  
\begin{equation}
    \arg\max_{d_i\in\mathcal{N}(q)}\{\mathcal{D}_{\mathrm{PQ}}\bigl(\mathrm{code}(\phi(q')),\,\mathrm{code}(\phi(d_i))\bigr)\} 
\end{equation}
where $\mathcal{N}(q) \subset \mathcal{D}$ denotes the set of images that are semantically and geographically relevant to $q$. The perturbation $\delta$ is subject to a perception budget constraint:
\begin{equation}
    \|\delta\|_{\infty} \le \epsilon
\end{equation}
\subsection{Feature Encoder}
The feature encoder $\phi$ maps an input image $q$ to $d$-dimensional $\ell_2$-normalized embedding. The architecture follows the standard design:
\begin{equation}
    \mathbf{f}=\phi(q)=\frac{\mathrm{FC}(\mathrm{Conv}(q))}{\left\|\mathrm{FC}(\mathrm{Conv}(q))\right\|_2}
\end{equation}
where $\mathrm{Conv}(\cdot)$ represents the hierarchical stack of convolutional layers responsible for extracting spatially-aware feature maps, and $\mathrm{FC}(\cdot)$ denotes the fully connected projection head that aggregates local features into a single global descriptor. The $\ell_2$ normalization ensures that cosine similarity between features serves as an effective proxy for retrieval relevance.

In the adversarial generation phase, $\phi$ serves as a deterministic differentiable mapping from pixel space to feature space, through which gradients can be back-propagated. Critically, the parameters of $\Phi$ are not updated during adversarial optimization, and only the input perturbation $\delta$ is learned.
\subsection{Soft Centroid Distribution and the Differentiability}
The fundamental obstacle in attacking PQ-based systems is the non-differentiability of the hard nearest-centroid assignment $b^{(l)}$. This argmin operation has zero gradient almost everywhere, making it impossible to propagate meaningful gradient signals back to the input perturbation $\delta$. To circumvent this, we adopt a soft centroid assignment strategy inspired by the product quantization network \cite{chen2022adversarial}. Specifically, for the $l$-th subspace, we compute the soft probability distribution $\mathbf{p}^{(l)}=\left(p_1^{(l)},\,p_2^{(l)},\,\cdots,\,p_K^{(l)}\right)$ over all $K$ centroids based on the cosine similarity between the sub-feature vector and each centroid:
\begin{equation}
    p_k^{(l)}=\frac{\exp\!\left(\tau\cdot\langle \hat{f}^{(l)},\,c_k^{(l)}\rangle\right)}{\sum_{k'=1}^{K}\exp\!\left(\tau\cdot\langle \hat{f}^{(l)},\,c_{k'}^{(l)}\rangle\right)},\quad k\in[K]
\end{equation}
where $\hat{f}^{(l)}$ denotes the (possibly perturbed) $l$-th sub-vector, $\langle \cdot,\cdot\rangle$ denotes the inner product, and $\tau>0$ is a temperature hyperparameter that controls the sharpness of the distribution. As $\tau\rightarrow\infty$, the soft distribution converges to the one-hot hard assignment; as $\tau\rightarrow0$, the distribution becomes uniform over all centroids. This formulation is fully differentiable with respect to $\hat{f}^{(l)}$, enabling end-to-end gradient flow from the distribution to the pixel space.

Motivated by the observation that the soft probability distribution $p^{(l)}$encodes the full geometric relationship between a feature sub-vector and the codebook, we identify two complementary strategies for adversarial distribution perturbation:
\subsubsection{Peak-Targeted Perturbation (PTP)} This strategy focuses on disrupting the dominant assignment by minimizing the probability assigned to the ground-truth centroid $b^{(l)}$. Using the cross-entropy loss with the one-hot ground-truth distribution $p^{(l)}_{gt}=onehot(b^{(l)})$ as a reference:
\begin{equation}
    \mathcal{L}_{\mathrm{PTP}}^{(l)}
=
-\sum_{k=1}^{K}\mathbf{1}\!\left[k=b^{(l)}\right]\cdot \log p_k^{(l)}
=
-\log p_{b^{(l)}}^{(l)}
\end{equation}

While PTP is conceptually straightforward, it carries a critical limitation: for symmetric retrieval settings, reducing the probability of the peak centroid may merely redistribute probability mass among adjacent centroids with similar geometric properties, producing only a marginal increase in quantization error.
\subsubsection{Distribution-Wide Perturbation (DWP)} DWP seeks to maximize the divergence between the adversarial centroid distribution and the clean distribution, disrupting the entire probabilistic structure. The attack is formulated as minimizing the cross-entropy between the clean soft distribution $p^{(l)}$ (computed from the clean query) and the adversarial distribution $\hat{p}_k^{(l)}$ (computed from the perturbed query):
\begin{equation}
    \mathcal{L}_{\mathrm{DWP}}^{(l)}
=
-\sum_{k=1}^{K} p_k^{(l)} \cdot \log \hat{p}_k^{(l)}
\end{equation}
It can be verified that minimizing $\mathcal{L}_{\mathrm{DWP}}^{(l)}$ with respect to $\hat{p}_k^{(l)}$ is equivalent to maximizing the KL divergence $\mathrm{KL}\!\left(\mathbf{p}^{(l)} \,\|\, \hat{\mathbf{p}}^{(l)}\right)$:
\begin{equation}
    \max_{\hat{p}^{(l)}} \mathrm{KL}\!\left(p^{(l)} \,\|\, \hat{p}^{(l)}\right)
=
\max_{\hat{p}^{(l)}} \sum_{k=1}^{K} p_k^{(l)} \log \frac{p_k^{(l)}}{\hat{p}_k^{(l)}}
=
\min_{\hat{p}^{(l)}} \mathcal{L}_{\mathrm{DWP}}^{(l)}
\end{equation}

This distributional formulation is particularly effective in asymmetric retrieval settings, where the similarity score between a query and a database entry is directly determined by the inner product structure of the adversarial distribution, making global distributional corruption more damaging than local peak attacks.
\subsection{Lightweight Product Quantization Network (LPQN)}

The core architectural unit of the LPQN is the Quantization Block, which processes the feature sub-vector for the $l$-th subspace and produces a soft quantization code. As shown in Figure \ref{fig3}. The Q-B consists of two coupled components:
\begin{figure}
    \centering
    \includegraphics[width=0.95\linewidth]{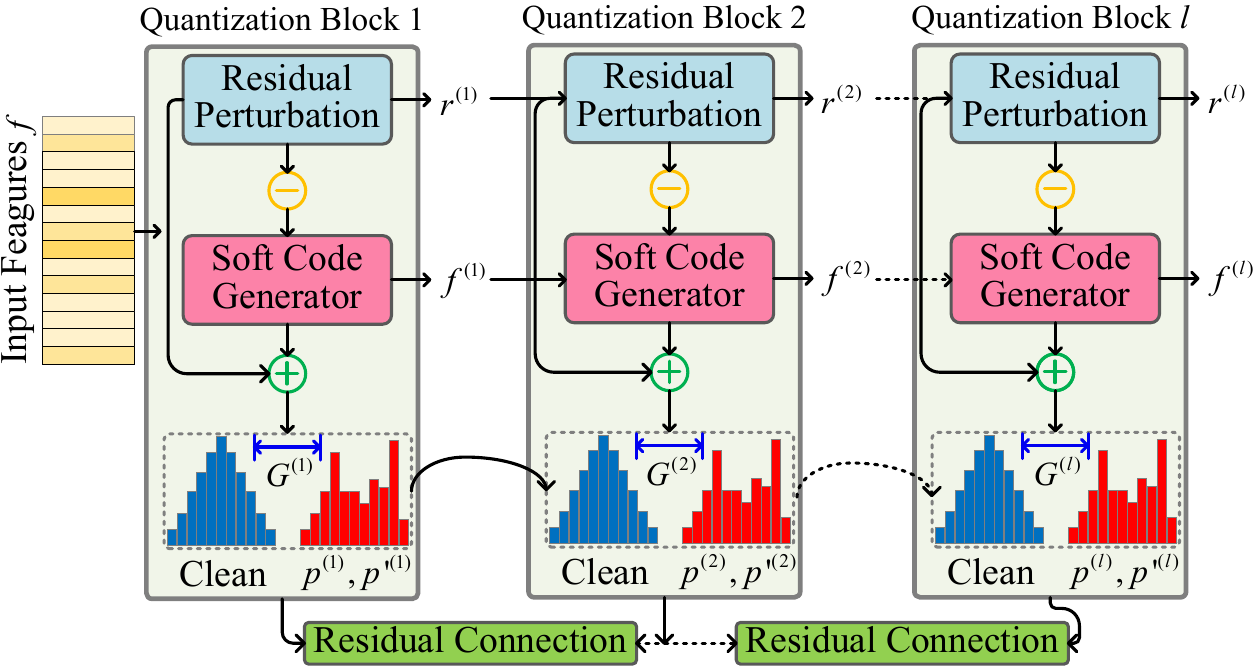}
    \caption{The proposed lightweight product quantization network (LPQN) with residual connection.}
    \label{fig3}
\end{figure}
\subsubsection{Residual Perturbation Module} Before computing the centroid assignment, the Q-B injects a learnable residual offset $r^{(l)} \in \mathbb{R}^{d/L}$ into the sub-feature vector:
\begin{equation}
    \tilde{f}^{(l)} = f^{(l)} + r^{(l)}
\end{equation}
where the residual $r^{(l)}$ is constrained by $\|r^{(l)}\|_{\infty} \le \epsilon_l$ to ensure that subspace-level perturbations remain bounded. This design decouples perturbation injection from the feature encoder, enabling targeted disruption of the encoding process without modifying the upstream feature extraction.
\subsubsection{Soft Code Generator} The perturbed sub-vector $\tilde{f}^{(l)}$ is then passed through the differentiable soft assignment layer to produce a soft code $q^{(l)} \in \Delta^{K-1}
$:
\begin{equation}
    q_k^{(l)}=\frac{\exp\!\left(\tau\cdot\langle \tilde{f}^{(l)},\,c_k^{(l)}\rangle\right)}{\sum_{k'=1}^{K}\exp\!\left(\tau\cdot\langle \tilde{f}^{(l)},\,c_{k'}^{(l)}\rangle\right)}
\end{equation}
where $\Delta^{K-1}$ denotes the $(K-1)$-dimensional probability simplex. The resulting Q-Code $\mathbf{q}^{(l)}
q(l)$ provides a differentiable, continuous relaxation of the discrete hard assignment code.
\subsubsection{Progressive Binary Code Aggregation} A key design feature of the LPQN is its progressive code construction mechanism, wherein quantization codes from successive Q-Bs are incrementally fused via additive accumulation $\oplus$. After processing the first $l$ subspaces, the aggregated code has length $l\cdot m$ bits:
\begin{equation}
    \mathrm{Code}_{\mathrm{agg}}^{(l)}=\bigoplus_{j=1}^{l} q^{(j)},\quad l=1,2,\ldots,L
\end{equation}

This progressive structure, illustrated in Figure \ref{fig2} as the chain $1m \rightarrow 2m \rightarrow \cdots \rightarrow Lm$, yields several important properties.
\subsubsection{Residual Skip Connections Between Q-B$_s$} As shown in Figure \ref{fig2}, consecutive Q-B$_s$ are connected via residual pathways. The input Q-B$_{l+1}$ to is not the raw feature sub-vector but rather the feature representation after applying the adversarial residual from Q-B$_l$:
\begin{equation}
    f^{(l+1)} = f^{(l)} + r^{(l+1)} \cdot \mathcal{G}^{(l)}
\end{equation}
where $\mathcal{G}^{(l)}$ represents a gating mechanism conditioned on the code disruption magnitude achieved by the preceding block. 
\begin{table*}[h]
\centering
\caption{Recall@1 performance comparison under different adversarial attacks}
\label{tab1}
\resizebox{\linewidth}{!}{
\begin{tabular}{||l||ccc||ccc||ccc||ccc||ccc||ccc||}
\hline
\multirow{3}{*}{Methods} & \multicolumn{9}{c||}{Pittsburgh250k} & \multicolumn{9}{c||}{Tokyo24/7} \\
\cline{2-19}
& \multicolumn{3}{c||}{AlexNet} & \multicolumn{3}{c||}{VGG16} & \multicolumn{3}{c||}{ResNet18} & \multicolumn{3}{c||}{AlexNet} & \multicolumn{3}{c||}{VGG16} & \multicolumn{3}{c||}{ResNet18} \\
\cline{2-19}
& 16 & 32 & 64 & 16 & 32 & 64 & 16 & 32 & 64 & 16 & 32 & 64 & 16 & 32 & 64 & 16 & 32 & 64 \\
\hline
Clean & 68.3 & 72.1 & 74.8 & 78.5 & 82.3 & 84.9 & 81.7 & 85.2 & 87.5 & 62.4 & 66.8 & 69.3 & 72.1 & 76.5 & 79.1 & 75.3 & 79.8 & 82.2 \\
FGSM \cite{goodfellow2014explaining} & 52.6 & 58.3 & 62.1 & 61.2 & 67.5 & 71.4 & 64.8 & 71.2 & 75.3 & 47.3 & 53.1 & 56.8 & 56.7 & 62.4 & 66.2 & 59.2 & 65.7 & 69.8 \\
PGD \cite{gupta2018cnn}& 38.5 & 45.7 & 50.8 & 47.3 & 55.8 & 61.2 & 51.2 & 59.6 & 65.4 & 34.1 & 41.2 & 46.1 & 42.8 & 50.3 & 55.9 & 46.5 & 54.2 & 60.2 \\
UAPR \cite{li2019universal}& 44.7 & 51.2 & 56.3 & 53.6 & 61.4 & 66.8 & 57.8 & 65.3 & 70.7 & 39.8 & 46.5 & 51.4 & 48.9 & 56.7 & 62.1 & 52.1 & 60.1 & 65.8 \\ \hline
\rowcolor{gray!15} LPQN-DWP & 28.4 & 36.8 & 42.5 & 35.7 & 45.2 & 52.1 & 39.6 & 49.8 & 57.2 & 24.6 & 32.9 & 38.7 & 31.2 & 40.6 & 47.3 & 34.8 & 44.7 & 52.1 \\
\rowcolor{gray!15} LPQN-PTP & 31.5 & 39.6 & 45.7 & 39.2 & 48.7 & 55.3 & 43.1 & 53.2 & 60.5 & 27.8 & 35.7 & 41.8 & 34.5 & 43.9 & 50.7 & 38.2 & 48.1 & 55.3 \\
\rowcolor{gray!15} LPQN-Hybrid & \textbf{25.7} & \textbf{34.2} & \textbf{40.1} & \textbf{32.8} & \textbf{42.6} & \textbf{49.5} & \textbf{36.9} & \textbf{47.1} & \textbf{54.8} & \textbf{22.1} & \textbf{30.4} & \textbf{36.2} & \textbf{28.6} & \textbf{38.2} & \textbf{45.1} & \textbf{32.3} & \textbf{42.3} & \textbf{49.7} \\
\hline
\end{tabular}
}
\end{table*}
\subsubsection{Two-Phase Adversarial Query Generation} In the forward pass, the benign query feature $f=\phi(q)$ is propagated through the LPQN, with each Q-B simultaneously computing the clean soft distribution $p^{(l)}$ and the adversarially perturbed distribution $\hat{p}^{(l)}$.
For each subspace $l$, we formulate the forward attack objective using the DWP strategy as the primary mechanism with PTP as an optional mode:
    \begin{equation}
        \mathcal{L}_{\mathrm{fwd}}^{(l)}
=
-\sum_{k=1}^{K} p_k^{(l)} \cdot \log \hat{p}_k^{(l)}
+\lambda \cdot \|r^{(l)}\|_2^2
    \end{equation}
where the first term drives the adversarial distribution away from the clean distribution by maximizing their KL divergence, and the second term regularizes the magnitude of the residual perturbation with weight $\ \lambda>0$ to prevent excessive subspace-level distortion.

The overall forward loss is computed by aggregating contributions from all $L$ subspaces:
\begin{equation}
    \mathcal{L}_{\mathrm{fwd}}=\sum_{l=1}^{L}\omega_l \cdot \mathcal{L}_{\mathrm{fwd}}^{(l)}
\end{equation}
where $\omega_l > 0$ are subspace-specific weighting coefficients that can be adapted to the relative importance of each subspace. In the default configuration, we set $\omega_l = 1/L$ for uniform weighting across subspaces.

While the forward pass optimizes the subspace-level residuals $\left\{r^{(l)}\right\}_{l=1}^{L}$ within the LPQN, the backward pass propagates the attack objective through the full computational graph. The backward optimization minimizes a composite loss that jointly targets both distributional disruption and retrieval performance degradation:
\begin{equation}
    \mathcal{L}_{\mathrm{bwd}}=\mathcal{L}_{\mathrm{fwd}}+\mu\cdot \mathcal{L}_{\mathrm{sep}}
\end{equation}
where $\mu > 0$ balances the two terms, and $\mathcal{L}_{\mathrm{sep}}$ is a feature separation loss that explicitly pushes the adversarial feature $\Phi(q')$ away from the clean feature $\phi(q)$ in the deep embedding space:
\begin{equation}
    \mathcal{L}_{\mathrm{sep}}=-\|\phi(q+\delta)-\phi(q)\|_2^2
\end{equation}

This dual objective ensures that the adversarial perturbation operates at both the distribution level and the feature level, achieving complementary and mutually reinforcing disruption mechanisms.
\begin{figure}[http]
    \centering
    \includegraphics[width=1\linewidth]{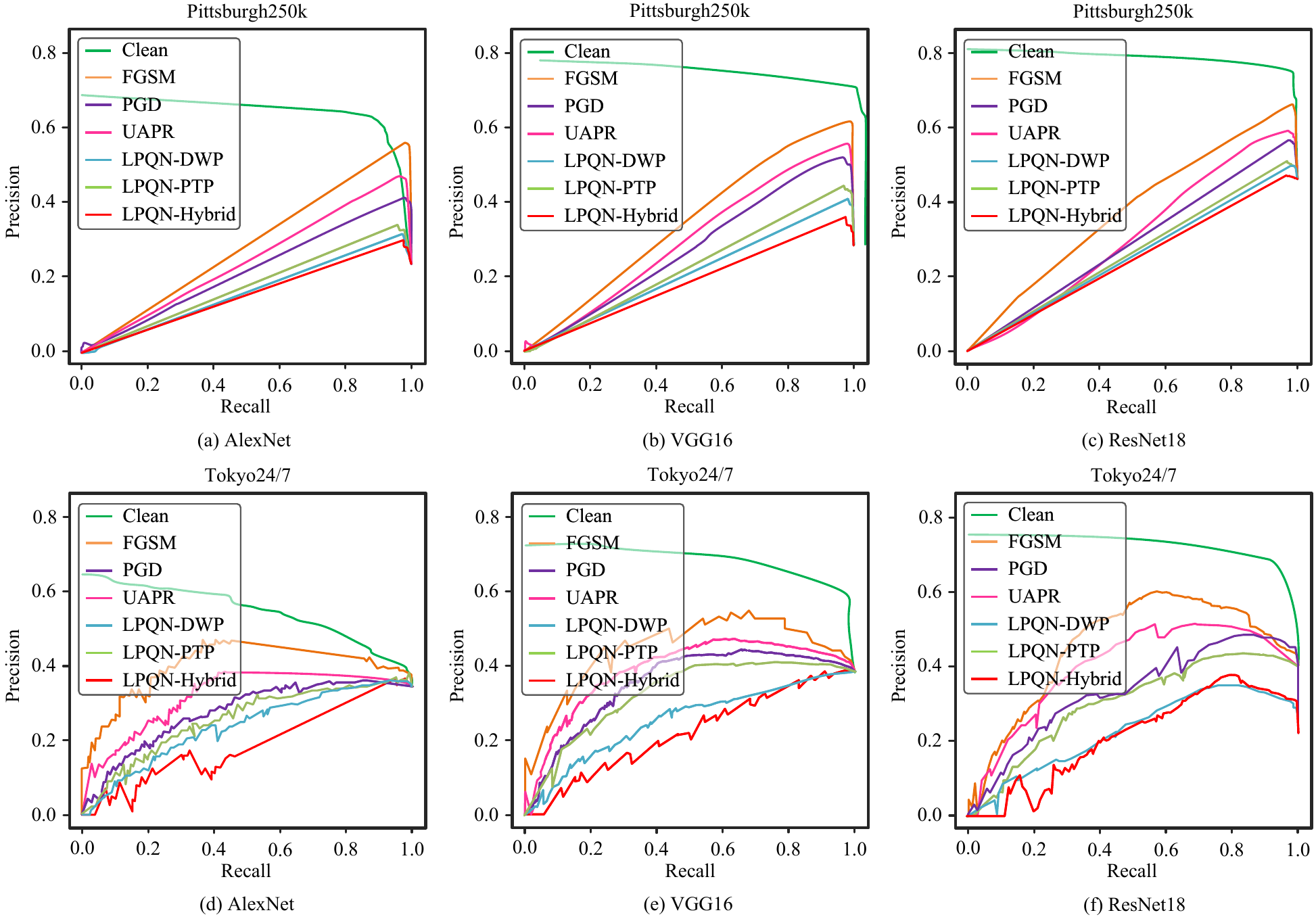}
    \caption{Precision-Recall curve for AlexNet, VGG16, and ResNet18 based on Pittsburgh250k and Tokyo24/7, respectively.}
    \label{fig4}
\end{figure}
\begin{figure}[htbp]
\label{fig5}
    \centering
    \includegraphics[width=0.23\textwidth]{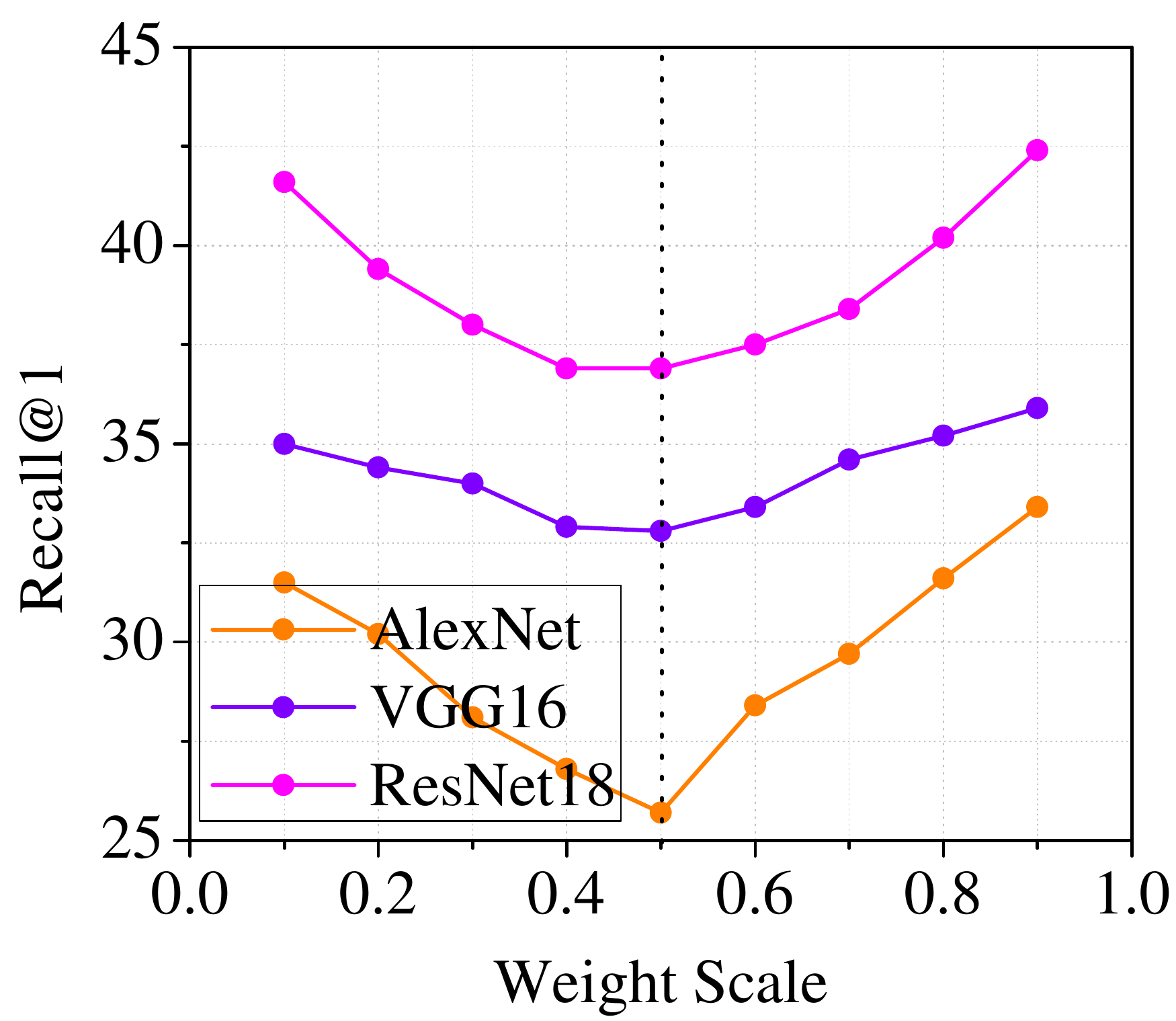}
    \hfill
    \includegraphics[width=0.23\textwidth]{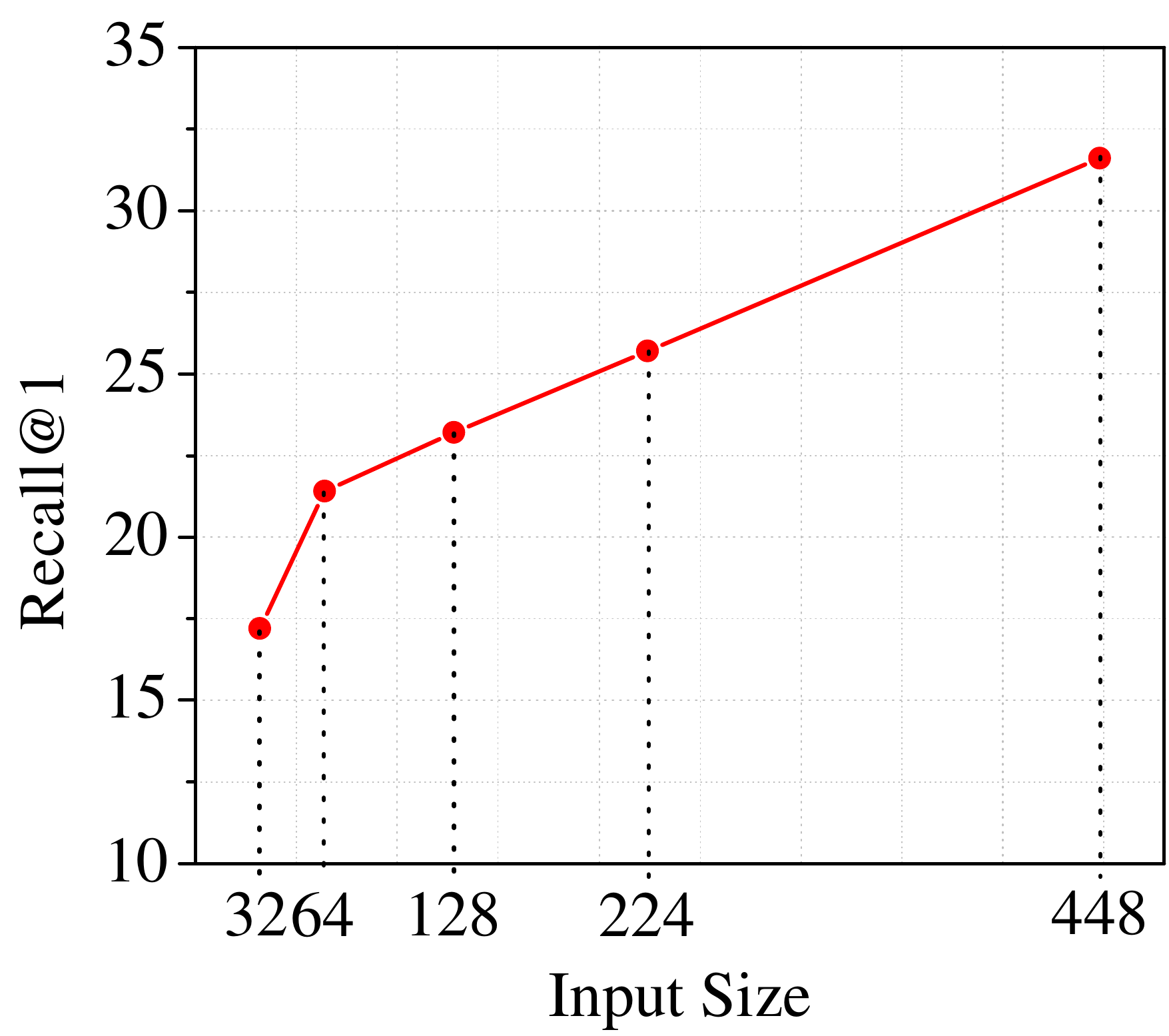}
    \caption{Ablation study on weight scale $(\mathcal{L}_{\mathrm{DWP}}/(\mathcal{L}_{\mathrm{DWP}} +  \mathcal{L}_{\mathrm{PTP}})$ and input size, including $32\times32$, $64\times64$, $128\times128$, $224\times224$, and $448\times448$.}
\end{figure}
\begin{figure*}
    \centering
    \includegraphics[width=0.75\linewidth]{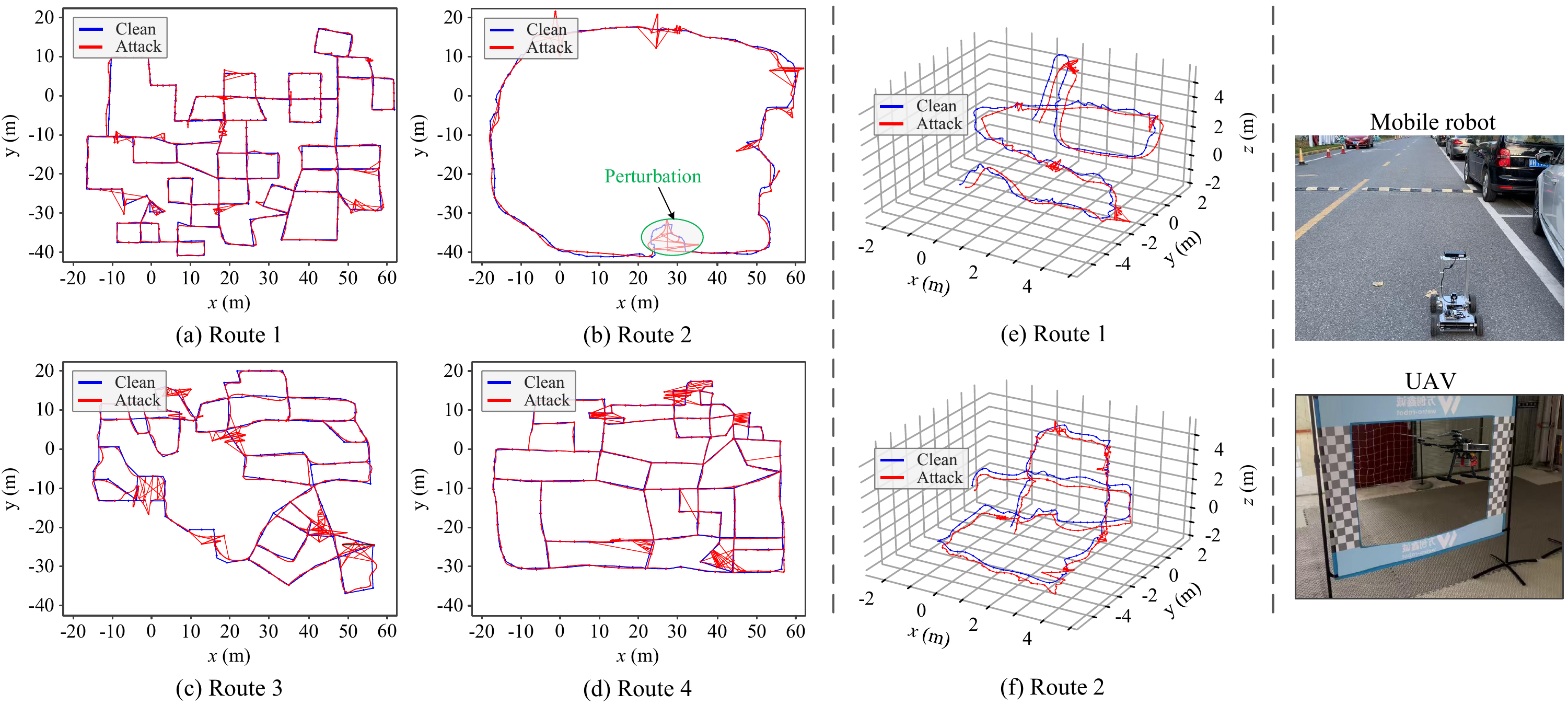}
    \caption{We conducted real-world localization experiments based on the LPQN-Hybrid strategy, using both a mobile robot and an UAV. Figure 6(a)-(d) show experiments based on the mobile robot platform, and (e)-(f) show experiments based on the UAV platform. The figures include two robot trajectories, representing the localization results without attack and with the proposed LPQN-Hybrid, respectively. The points in the figures indicate the localizations at which the robots successfully matched the database locations.}
    \label{fig6}
    \end{figure*}
\section{Experiments}
\subsection{Implementation Details}
Our experiments were conducted on the Pittsburgh250k (250,000 images) and Tokyo24/7 (76,000 images) benchmarks, using ResNet-50 with NetVLAD aggregation to generate 4096-dimensional $\ell_2$-normalized descriptors from $224 \times 224$ input images. The product quantization layer decomposed features into $L=8$ subspaces (512 dimensions each) with $K=256$ centroids per codebook learned via $k$-means (50 iterations, convergence threshold $<10^{-4}$), yielding 64-bit codes with softmax temperature $\tau=10.0$. The LPQN consisted of eight quantization blocks with adaptive residual gating ($\gamma=0.5$), optimized via 100 Projected Gradient Descent (PGD) \cite{kurakin2018adversarial} iterations with step size $\alpha=1/255$, global perturbation budget $\epsilon=8/255$, and subspace-level constraint $\epsilon_l=0.1$. Loss hyperparameters were configured as: regularization weight $\lambda=0.01$, feature separation weight $\mu=0.5$, and uniform subspace weighting $\omega_l=1/L$. We evaluated three attack variants: 1) LPQN-DWP (Distribution-Wide Perturbation), maximizing KL divergence across all centroids in the soft assignment distribution (Eq. 9); 2) LPQN-PTP (Peak-Targeted Perturbation), minimizing the probability of the ground-truth centroid (Eq. 8); and 3) LPQN-Hybrid, combining both objectives with equal weighting ($0.5 \times \mathcal{L}_{\mathrm{DWP}} + 0.5 \times \mathcal{L}_{\mathrm{PTP}}$). These were compared against baseline adversarial methods: Fast Gradient Sign Method (FGSM) \cite{goodfellow2014explaining}, Projected Gradient Descent (PGD) \cite{gupta2018cnn}, and Universal Adversarial Perturbations for Retrieval (UAPR) \cite{li2019universal}. Performance was assessed using $\mathrm{Recall}@K$ for $K \in \{1,5,10,20\}$ with a 25-meter GPS tolerance for correct retrievals.
\subsection{Performance Evaluation of Attack Effects}
Table \ref{tab1} demonstrates critical vulnerabilities in product quantization-based localization systems across various architectures, descriptor lengths, and benchmarks. The LPQN-Hybrid method demonstrates superior attack effectiveness, reducing Recall@1 from 85.2\% to 36.9\% on Pittsburgh250k with ResNet 18 and 32-bit codes, compared to 59.6\% under PGD and 71.2\% under FGSM. Even with extended 64-bit codes, which offer stronger representational capacity, the attack still causes substantial degradation to 54.8\%. This 15 to 20 percentage point superiority over baseline methods validates that explicitly targeting centroid assignment distributions outperforms pixel-level perturbations, with the hybrid strategy, which combines distribution-wide and peak-targeted disruptions, maximizes retrieval failure.

Attack effectiveness exhibits critical dependency on descriptor length across all methods. Under LPQN-Hybrid, 16-bit codes suffer 68\% degradation compared to 60\% for 32-bit and 52\% for 64-bit codes, revealing that while longer codes provide improved robustness through more stable centroid boundaries, they remain fundamentally vulnerable to quantization-space attacks. 

As shown in Figure 4, to ensure fair comparison and facilitate clearer visualization, all precision-recall curves were recalculated using 16-bit quantization. We observed that, unlike the unattacked network (clean), the curves of the attacked network exhibited a monotonically increasing trend. Furthermore, at the top of the retrieval region, the difference between the curves and the original curves increased significantly, leading to a decrease in precision and recall. This indicates that the proposed attack strategy destroys the semantic information within the feature space, effectively pushing the features of the adversarial query away from the original relevant features stored in the database.

\subsection{Ablation Study}
We conducted ablation studies to evaluate the impact of key parameters on the localization system. Specifically, we examined the effects of the weight scale and input image size on attack effectiveness. As shown in Figure 5, the attack on LPQN is most effective when the weight ratios of DWP and PTP in the hybrid attack are equal; however, the attack weakens as the ratio deviates from this balance. Additionally, the attack's effectiveness decreases as the image size increases, primarily because larger images generate more feature points, which fragment the attack and reduce its overall impact.
\subsection{Generalization Evaluation}
We also tested the generalization effect of the proposed attack method by injecting LPQN-Hybrid perturbations into the feature space of the mainstream visual localization methods. We also tested the generalization effect of the proposed attack method by injecting LPQN-Hybrid perturbations into the feature space of the mainstream visual localization methods. To maintain the original state of the model, we merely add our LPQN between the feature encoding and feature clustering of these models, and immediately insert a reverse LPQN to ensure that the model maintains the original output unchanged. 

Based on the results presented in Table 2, the LPQN-Hybrid attack consistently and markedly degrades the retrieval performance of all evaluated visual localization methods across the three datasets (Pitts30k, MSLS-V, and MSLS-C). The attack is injected into the feature space between feature encoding and clustering, while a reverse LPQN keeps the original model output unchanged, a setting that isolates the pure effect of the proposed perturbation.

Under this setup, every method suffers a substantial drop in Recall@1 and Recall@5 after the attack. Notably, even the most competitive baselines exhibit clear performance deterioration, while earlier or less robust models approach near-failure levels on the more challenging datasets, such as MSLS-C. These widespread and pronounced degradations demonstrate that LPQN-Hybrid can effectively disrupt feature representations across diverse visual localization architectures.
\begin{table}[htbp]
\centering
\caption{Generalization evaluation of LPQN-Hybrid attacks on additional visual place recognition benchmarks. The data in the \textcolor{gray}{(/)} represents the Recall@1/5 rate after the attack.}
\resizebox{\linewidth}{!}{
\begin{tabular}{||c||c||c||c||}
\hline
 & Pitts30k        & MSLS-V       & MSLS-C \\ \cline{2-4} 
Method  & Recall@1/5             & Recall//5      & Recall@1/5       \\ \hline
NetVLAD  & 81.9/91.2 \textcolor{gray}{(60.8/72.6)}  & 53.1/66.5 \textcolor{gray}{(44.6/50.8)} & 35.1/47.4 \textcolor{gray}{(22.8/32.4)}\\ 
CosPlace   & 88.5/94.5 \textcolor{gray}{(68.8/85.1)}  & 82.8/89.7 \textcolor{gray}{(70.7/80.2)}& 61.4/72.0 \textcolor{gray}{(52.4/64.3)}\\ 
MixVPR  & 91.5/95.5 \textcolor{gray}{(72.5/79.6)}  & 88.2/93.1 \textcolor{gray}{(71.8/84.5)}& 64.0/75.9 \textcolor{gray}{(50.8/67.2)}\\ 
EigenPlaces  & 92.5/96.8 \textcolor{gray}{(78.4/80.3)}  & 89.1/93.8 \textcolor{gray}{(76.4/80.9)}& 67.4/77.1 \textcolor{gray}{(50.3/64.7)}\\ 
SelaVPR  & 92.8/96.8 \textcolor{gray}{(77.8/82.6)} & 90.8/96.4 \textcolor{gray}{(79.2/85.4)}& 73.5/87.5 \textcolor{gray}{(64.8/75.7)}\\ 
CricaVPR  & 94.9/97.3 \textcolor{gray}{(78.4/82.8)} & 90.0/95.4 \textcolor{gray}{(78.6/86.4)}& 69.0/82.1 \textcolor{gray}{(49.1/66.7)}\\ 
TransVPR  & 89.0/94.9 \textcolor{gray}{(70.0/79.8)}& 86.8/91.2 \textcolor{gray}{(80.2/84.5)}& 63.9/74.0 \textcolor{gray}{(52.3/66.4)}\\ 
SALAD & 92.4/96.3 \textcolor{gray}{(80.5/87.6)}  & 92.2/96.2 \textcolor{gray}{(84.3/88.5)}& 75.0/88.8 \textcolor{gray}{(66.7/80.1)}\\ 
FoL  & 94.5/97.4 \textcolor{gray}{(77.5/85.4)}  & 93.5/96.9 \textcolor{gray}{(82.6/88.7)}& 80.0/90.9 \textcolor{gray}{(70.3/84.3)}\\ 
Pair-VPR   & 95.4/97.5 \textcolor{gray}{(79.6/85.4)}  & 95.4/97.3 \textcolor{gray}{(82.4/89.8)}& 81.7/90.2 \textcolor{gray}{(72.5/81.6)}\\ 
$A^2$GC  & 94.6/98.8 \textcolor{gray}{(85.6/89.9)}& 93.0/97.3 \textcolor{gray}{(84.4/89.9)}& 79.7/90.2 \textcolor{gray}{(72.2/85.7)}\\ \hline
\end{tabular}
}
\label{table1}
\end{table}
\subsection{Real-World Test}
 Real-world validation was performed on a mobile robot and UAV platform equipped with an NVIDIA Jetson Orin NX and an RGB-D camera, operating autonomously in an outdoor environment using ROS Noetic. The robots matched live camera frames captured at 10 Hz against a pre-built reference database of 500 images, with adversarial perturbations generated offline on an NVIDIA RTX 3090 GPU. It was injected into the localization pipeline at 1 Hz during navigation tasks. Adversarial perturbations were generated offline using the datasets by training an LQPN and subsequently deployed to the computational unit of a physical robot for online testing. To accelerate the testing process, we employed five convolutional layers followed by a fully connected layer as the feature encoder, whereas the LQPN uses 8 modules with 8-bit outputs.
 
 In the experiment, we used the sparse locations of each route, collected beforehand, as a reference database to match the query locations during the real-time driving process of the physical robots. The localization results obtained without attacks were used as the baseline. The points shown in Figure \ref{fig6} indicate that the current query successfully matched the database reference locations. Compared with the clean test (baseline), the attack-based test produced multiple deviations and perturbations, demonstrating that the attack significantly impacted the localization system.
 
\begin{figure}[http]
    \centering
    \includegraphics[width=0.99\linewidth]{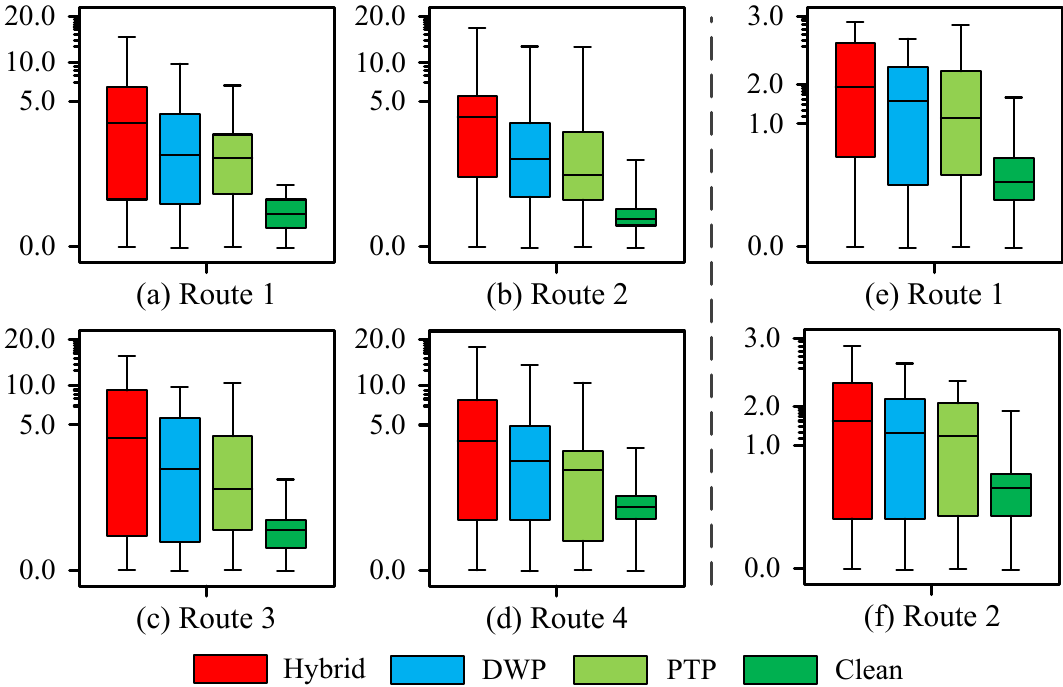}
    \caption{Localization error with the proposed four LPQN-based attack strategies. (a)-(d): attacking with a mobile robot platform. (e)-(f): attacking with a UAV platform.}
    \label{fig7}
\end{figure}

 Furthermore, we compared the test results for the six routes against GPS-based ground truth to compute the localization error for each attack strategy. As shown in Figure \ref{fig7}, the localization error was the largest under LPQN-Hybrid and the smallest under no attack (clean), indicating that the feature space attack based on LPQN is very effective.
\subsection{Deployment Cost Analysis}
The runtime analysis in Table 3 indicates that the proposed attack incurs minimal computational overhead on the localization pipeline. Feature extraction and retrieval dominate the latency but still require relatively little processing time due to the use of simple convolutional layers and a fast matching method. Meanwhile, the adversarial perturbation module operates at a low frequency (1 Hz), adding only a marginal additional cost. Even under attack conditions, the system maintains real-time performance (10 Hz), demonstrating the practicality of the proposed method for real-world robotic deployment.
\begin{table}
\caption{Deployment cost analysis for robot localization test with clean and attack states.}
\resizebox{\linewidth}{!}{
\begin{tabular}{||c||c||c||c||}
\hline
Module                        & Operation                                & Frequency                & Latency           \\ \hline
Image   Acquisition           & Frame   capture                          & \multirow{5}{*}{10   Hz} & 7   ms            \\ 
Feature   Extraction          & Forward   pass                           &                          & 10   ms           \\ 
Feature   Normalization       & Normalization                       &                          & \textless{}1   ms \\ 
PQ   Encoding                 &  Quantization                  &                          & 7   ms            \\ 
Database   Retrieval          & Search                                   &                          & 8   ms           \\ 
Adversarial   Perturbation    & LPQN   inference                         & 1   Hz                   & 23   ms           \\ 
Attack   Generation & Synthesis & Offline                  & 0.05   s/sample   \\ \hline
\rowcolor{gray!15} Total   (Clean)               & -            & - &     32 ms         \\ 
\rowcolor{gray!15} Total   (Attack)              & -    &        -                  &    55 ms         \\ \hline
\end{tabular}
}
\label{tab2}
\end{table}
\section{Conclusions}
We propose a Lightweight Product Quantization Network (LPQN) to generate adversarial queries by perturbing soft centroid distributions. The two-phase optimization framework enables joint distributional and feature-level disruption. At the same time, the combination of distribution-wide perturbation (DWP) and peak-targeted perturbation (PTP) provides a unified attack formulation across local and global structures. Extensive experiments demonstrate that the proposed method consistently achieves stronger performance degradation than existing approaches across datasets and architectures. The results further indicate that increasing code length does not fundamentally improve robustness, suggesting that the vulnerability originates from the quantization mechanism itself. Real-world experiments confirm that such attacks can induce significant localization errors in robotic systems.

\bibliography{aaai2026}

\end{document}